%% file: m5508.tex
\documentclass{ecai} 

\usepackage{latexsym}
\usepackage{amssymb}
\usepackage{amsmath}
\usepackage{amsthm}
\usepackage{booktabs}
\usepackage{enumitem}
\usepackage{graphicx}
\usepackage{color}
\usepackage{multirow} 

\usepackage{algorithm}
\usepackage{algorithmic}

\usepackage{tabularx}
\usepackage{array}
\usepackage{makecell}
\usepackage{amssymb}
\usepackage{mathrsfs}
\usepackage{amsmath}
\usepackage{makecell}
\usepackage{float}
\usepackage{subcaption}
\usepackage{enumitem}
\usepackage{soul, color, xcolor}
\usepackage{arydshln}
\usepackage{hyperref}
\hypersetup{hidelinks,
	colorlinks=true,
	allcolors=blue,
	pdfstartview=Fit,
	breaklinks=true}

\newcommand{\BibTeX}{B\kern-.05em{\sc i\kern-.025em b}\kern-.08em\TeX}

\begin{document}


\begin{frontmatter}


\paperid{5508} 


\title{CRED-SQL: Enhancing Real-world Large Scale Database Text-to-SQL Parsing through Cluster Retrieval and Execution Description}


\author[A,B]{\fnms{Shaoming}~\snm{Duan}
\footnote{Equal contribution.}}
\author[A]{\fnms{Zirui}~\snm{Wang}
\footnotemark}
\author[A,B]{\fnms{Chuanyi}~\snm{Liu}\thanks{Corresponding Author. Email: liuchuanyi@hit.edu.cn}} 
\author[A]{\fnms{Zhibin}~\snm{Zhu}}
\author[A,C]{\fnms{Yuhao}~\snm{Zhang}}
\author[A,B]{\fnms{Peiyi}~\snm{Han}}
\author[A,D]{\fnms{Liang}~\snm{Yan}}
\author[E]{\fnms{Zewu}~\snm{Peng}}

\address[A]{Harbin Institute of Technology, Shenzhen, China}
\address[B]{Pengcheng Laboratory, Shenzhen, China}
\address[C]{Mindflow.ai}
\address[D]{Inspur Cloud Information Technology Co., Ltd, Jinan 250101, China}
\address[E]{Guangdong Power Grid Co., Ltd, China}


\begin{abstract}
Recent advances in large language models (LLMs) have significantly improved the accuracy of Text-to-SQL systems. However, a critical challenge remains: the semantic mismatch between natural language questions (NLQs) and their corresponding SQL queries. This issue is exacerbated in large-scale databases, where semantically similar attributes hinder schema linking and semantic drift during SQL generation, ultimately reducing model accuracy. To address these challenges, we introduce CRED-SQL, a framework designed for large-scale databases that integrates Cluster Retrieval and Execution Description. CRED-SQL first performs cluster-based large-scale schema retrieval to pinpoint the tables and columns most relevant to a given NLQ, alleviating schema mismatch. It then introduces an intermediate natural language representation—Execution Description Language (EDL)—to bridge the gap between NLQs and SQL. This reformulation decomposes the task into two stages: Text-to-EDL and EDL-to-SQL, leveraging LLMs’ strong general reasoning capabilities while reducing semantic deviation. Extensive experiments on two large-scale, cross-domain benchmarks—SpiderUnion and BirdUnion—demonstrate that CRED-SQL achieves new state-of-the-art (SOTA) performance, validating its effectiveness and scalability. Our code is available at \href{https://github.com/smduan/CRED-SQL.git}{https://github.com/smduan/CRED-SQL.git}

\end{abstract}

\end{frontmatter}

\input{1.introduction}

\input{2.Related_Works}

\input{3.Method}

\input{4.Experiment}
\input{Conclusion}



\begin{ack}

This study is supported by the National Key Research and Development Program of China under Grant 2023YFB3106504, Guangdong Provincial Key Laboratory of Novel Security Intelligence Technologies under Grant 2022B1212010005, the China Postdoctoral Science Foundation under Grant Number 2024M751555, the Major Key Project of PCL under Grant PCL2024A04, Shenzhen Science and Technology Program under Grant ZDSYS20210623091809029 and RCBS20221008093131089, the project of Guangdong Power Grid Co., Ltd. under Grant 037800KC23090005 and GDKJXM20231042.
\end{ack}



\bibliography{mybibfile}
\input{Appendix}

\end{document}

%% file: 1.introduction.tex
\section{Introduction}

Text-to-SQL, which translates natural language questions into SQL queries, significantly lowers the barrier for both lay and expert users in interacting with databases \cite{gu2023few,katsogiannis2023survey,li2023resdsql,li2024codes}. In recent years, advancements in large language models (LLMs) have greatly enhanced the performance of Text-to-SQL solutions \cite{gao2024text, li2024can, pourreza2024din}. Current LLM-powered approaches \cite{gao2024text, pourreza2024din, wang2023mac} improve SQL generation accuracy by decomposing complex Text-to-SQL tasks into simpler sub-tasks, such as schema retrieval and SQL generation with subsequent correction. These methods have achieved state-of-the-art (SOTA) results on prominent benchmarks like Spider \cite{yu2018spider} and Bird \cite{li2024can}. However, a fundamental challenge persists: the semantic mismatch \cite{eyal2023semantic,gan2021natural, pourreza2024din} between natural language questions (NLQs) and their corresponding SQL queries, which manifests primarily in two areas—schema mismatch and semantic deviation during SQL generation.

\begin{figure}[t]
  \centering
  \includegraphics[width=\linewidth]{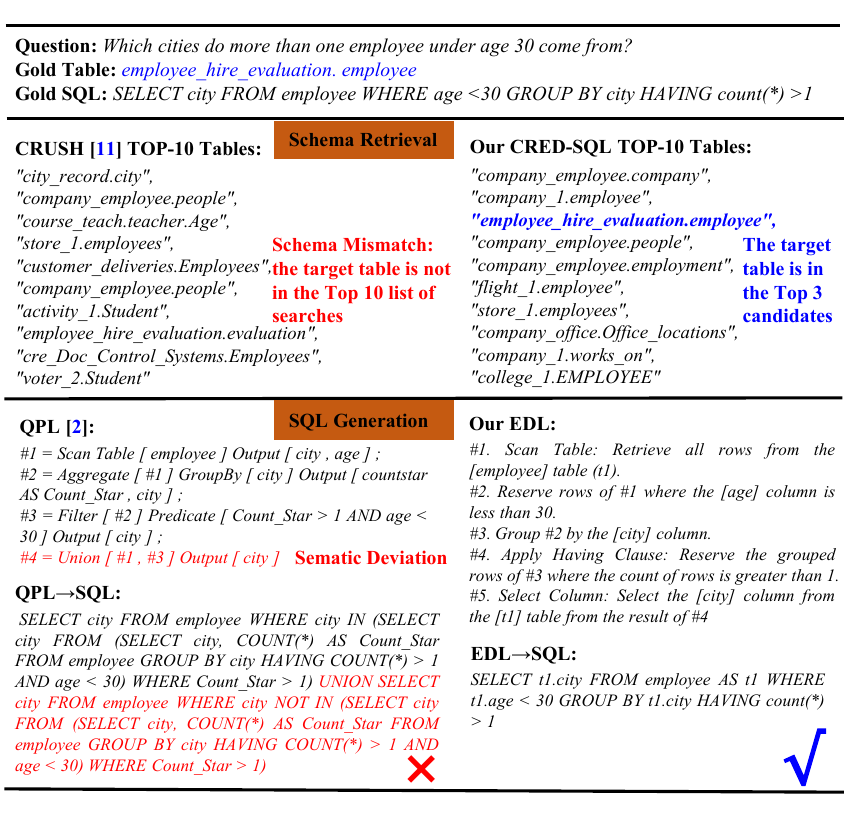}
  \caption{We illustrate an instance of schema mismatch during the schema retrieval phase and semantic deviation during SQL generation using the SpiderUnion dataset. Specifically, the SQL generation process involves the $employee\_hire\_evaluation$ database schema from Spider, which contains the target table.}
  \label{fig:example}
  \vspace{0.5cm}
\end{figure}

\textbf{Schema mismatch}. In large-scale, real-world databases, models often struggle to identify the correct tables and columns, leading to erroneous query generation.  To address this, many recent approaches \cite{pourreza2024din,talaei2024chess,  zhang2024finsql} perform schema retrieval by extracting a task-specific sub-schema before SQL generation. This technique is effective on benchmarks such as Spider \cite{yu2018spider} and Bird \cite{li2024can}, where databases are relatively small (e.g., Spider averages just 5.3 tables and 28.1 columns per database), allowing the entire schema to be embedded directly into the prompt.  However, in practical scenarios, schemas may contain thousands of tables and hundreds of columns—far exceeding the input limits of LLM prompts. Consequently, most studies \cite{kothyari2023crush4sql, muennighoff2022sgpt, zhang2024multi} first retrieve a high-recall subset of the schema to guide SQL generation. Unfortunately, when multiple tables and columns are semantically similar, existing retrieval techniques often fail to isolate the truly relevant elements, leading to retrieval errors. For example in figure \ref{fig:example}, the NLQ is: \textit{"Which cities do more than one employee under age 30 come from?" } The correct schema is the \textit{employee} table in the \textit{employee\_hire\_evaluation} database. However, due to the presence of numerous table and column names in the database that are semantically similar to keywords such as \textit{"cities", "employee"} and \textit{"age"} from the question, CRUSH \cite{kothyari2023crush4sql} ranks other tables as more relevant. As a result, the target table is not included among the top ten retrieved tables; instead, CRUSH returns tables associated with cities and people.

\textbf{Semantic deviation in SQL generation.} Directly mapping a natural language question (NLQ) to an SQL query involves bridging a substantial semantic gap. To mitigate this challenge, many studies \cite{eyal2023semantic, gan2021natural, guo2019towards} introduce intermediate SQL-like representations that are later translated into executable SQL. QPL \cite{eyal2023semantic}, for instance, enforces strict symbolic, structural, and database-semantic fidelity, which constrains the natural language generation capabilities of LLMs and limits their flexibility—particularly in handling complex operations such as numerical reasoning. As shown in Figure~\ref{fig:example}, given the NLQ: \textit{"Which cities do more than one employee under age 30 come from?"}, QPL is first generated using a fine-tuned LLM, and then the final SQL query is synthesized from the generated QPL. However, due to the limited ability of QPL to resolve semantic deviation between the NLQ and SQL, an error occurs in the fourth step: an incorrect \texttt{UNION} operation is introduced, deviating from the intended meaning of the question. This leads to erroneous SQL generation in subsequent steps.

To address these challenges, we propose CRED-SQL, a Text-to-SQL framework designed for large-scale databases that tackles both schema mismatch and semantic deviation through \textbf{C}luster \textbf{R}etrieval and \textbf{E}xecution \textbf{D}escription. Specifically, CRED-SQL introduces a cluster-based large-scale schema retrieval (CLSR) method to mitigate the impact of semantically similar attributes during schema retrieval. This method clusters tables and columns based on semantic similarity and applies a dynamic attribute-weighting strategy at query time: attributes most relevant to the NLQ are assigned higher weights, while irrelevant yet similar attributes are down-weighted, significantly improving schema selection. For SQL generation, we introduce Execution Description Language (EDL), a natural language representation that describes the intended SQL execution. By reformulating the Text-to-SQL task into two subtasks, Text-to-EDL and EDL-to-SQL, we better leverage the strengths of LLMs in general reasoning while reducing semantic deviation. To support this approach, we construct two new benchmarks by converting the Spider and Bird datasets into EDL descriptions and fine-tune an open-source LLM (Qwen2.5-Coder-32B) for the Text-to-EDL task. This fine-tuning substantially improves the model's ability to generate accurate EDLs from NLQs, and consequently, SQL queries from EDLs. Extensive experiments conducted on two large-scale, cross-domain datasets—SpiderUnion and BirdUnion—demonstrate that CRED-SQL surpasses existing SOTA baselines, validating its effectiveness and scalability.

The main contributions of this paper are as follows:

\begin{enumerate}[itemsep=2pt,topsep=-10pt,parsep=0pt]

\item We propose CRED-SQL, a Text-to-SQL framework for large-scale databases that addresses both schema mismatch and semantic deviation. CRED-SQL can be seamlessly integrated into existing Text-to-SQL systems employing reflection strategies, without any architectural modifications. 

\item We introduce a novel schema retrieval method for large-scale databases based on semantic similarity clustering. This method effectively mitigates the impact of a large number of semantically similar attributes on schema retrieval, improving retrieval accuracy.

\item We design Execution Description Language (EDL), a natural-language-based intermediate representation, and release two corresponding Text-to-EDL benchmarks built from the Spider and Bird datasets. Fine-tuning the open-source Qwen2.5-Coder-32B on these benchmarks yields higher Text-to-EDL accuracy than closed-source model such as GPT-4o.

\item We conduct comprehensive experiments on the two cross-domain datasets, SpiderUnion and BirdUnion, demonstrating that CRED-SQL achieves superior execution accuracy compared to SOTA methods powered by closed-source LLMs. We believe this framework offers a promising direction for advancing real-world Text-to-SQL applications. 
\end{enumerate}

%% file: 2.Related_Works.tex
\section{Related Works}
\subsection{Schema Retrieval for Large-scale Databases}

Existing schema retrieval methods \citep{ gao2024text, pourreza2024din,talaei2024chess, wang2023mac} primarily rely on LLMs by encoding the database schema and NLQ into prompts. DIN-SQL \cite{pourreza2024din} enhances this process with a Chain-of-Thought (CoT) \cite{wei2022chain} prompting strategy, guiding the model through step-by-step reasoning. DAIL-SQL \cite{gao2024text} improves performance via semantic similarity-based few-shot prompting. CHESS \cite{talaei2024chess} further introduces a hierarchical retrieval framework using model-generated keywords, LSH indexing, and vector databases to improve precision. These methods perform well on benchmarks such as Spider \cite{yu2018spider} and Bird \cite{li2024can}, but struggle with real-world databases that contain thousands of tables and columns—exceeding LLMs' context limits.

To address this, CRUSH \cite{kothyari2023crush4sql} proposes generating a minimal hallucinated schema via LLM, followed by dense passage retrieval (DPR) to find the most similar actual schema. MURRE \cite{zhang2024multi} enhances retrieval by mitigating issues from irrelevant or domain-mismatched entities through multi-hop DPR and question rewriting. Despite these advances, retrieval remains challenging in databases with many semantically similar attributes. Our method, CRED-SQL, tackles this by clustering attributes based on semantic similarity and down-weighting those in large clusters, which often represent generic or common fields. This reduces semantic interference and improves table retrieval accuracy in complex, real-world environments.

\subsection{Text-to-SQL with Intermediate Representations}

Intermediate representations (IRs) are used to simplify the conversion from natural language to SQL by capturing the semantics of queries in structured but flexible formats, avoiding SQL’s rigid syntax. SemQL \cite{lee1999semql}, a SQL-like language without the \texttt{FROM} clause, enables querying across autonomous databases. IRNet \cite{guo2019towards} synthesizes SemQL using a grammar-based model with a tree structure, omitting clauses like \texttt{GROUP BY} and \texttt{WHERE}, and then maps SemQL to SQL using domain knowledge. NatSQL \cite{gan2021natural} retains core SQL clauses (\texttt{SELECT, WHERE, ORDER BY}) while aligning its syntax with natural language, offering broader SQL compatibility than SemQL. However, both SemQL and NatSQL have limited support for set operations (e.g., \texttt{INTERSECT}) and nested queries, constraining their applicability to complex SQL tasks. QDMR \cite{zhang2011qdmr} decomposes complex queries into sequential sub-questions, aligning them with database operations. Though schema-agnostic and easy to annotate, QDMR suffers from error propagation across steps and lacks explicit schema handling. QPL \cite{eyal2023semantic} proposes a modular, semantically-driven IR that structures queries as layered sub-plans. It improves user interpretability and performance on datasets like Spider, but its limited syntactic coverage (e.g., no arithmetic operations) and strict structural format hinder LLM compatibility. In contrast, our CRED-SQL introduce Execution Description Language (EDL), a natural language-based IR that explicitly describes the intended SQL execution. By decoupling the Text-to-SQL task into Text-to-EDL and EDL-to-SQL, EDL leverages LLMs' reasoning strengths while reducing semantic mismatch and enhancing adaptability to complex queries.

%% file: 3.Method.tex
\section{Method}
\begin{figure*}
  \centering
  \includegraphics[width=\linewidth]{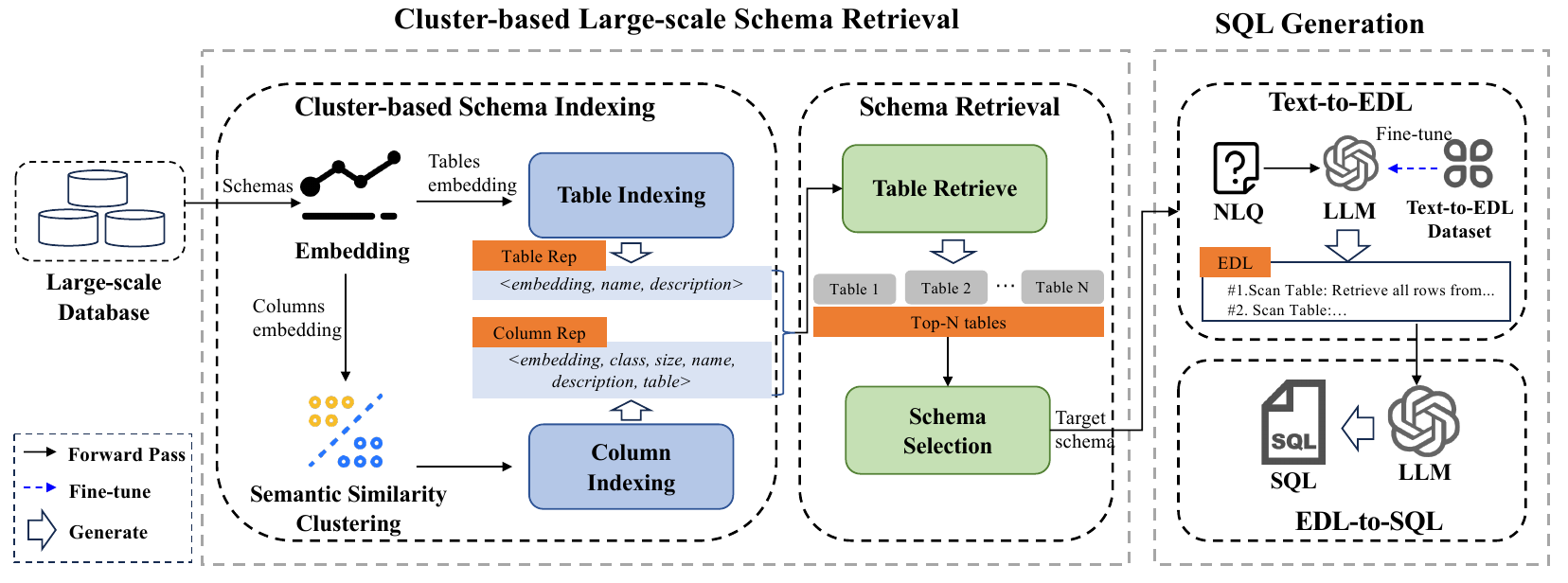}
  \caption{The overview of our CRED-SQL framework, which comprises two modules: (i) Cluster-based Large-scale Schema Retrieval, responsible for selecting relevant tables and columns from a large-scale database, and (ii) EDL-based SQL Generation, which translates the NLQ into an EDL representation and subsequently into executable SQL based on the retrieved schema.}
  \label{fig:cred-sql}
\end{figure*}

\subsection{Problem Definition}

In the Text-to-SQL task, given a set of natural language questions $Q=\{q_1,q_2,\dots,q_{|Q|}\}$ and a large-scale database schema $D=\langle T, C \rangle$—where $T=\{t_1,t_2,\dots,t_{|T|}\}$ denotes the set of tables and $C=\{c_1,c_2,\dots,c_{|C|}\}$ denotes the set of columns—the goal is to generate a corresponding SQL query $s_i$ for each question $q_i$ using a LLM, $s_{i}=\mathcal{M}_{NLQ \rightarrow SQL}(D, q_{i})$. However, in large-scale databases, the full schema $D$ often exceeds the context window of LLMs. To address this, prior work \cite{kothyari2023crush4sql} first retrieves a relevant schema subset $d_i$ from $D$:
\begin{equation}
  d_{i} = f(D, q_{i})
\end{equation}
\begin{equation}
s_{i}=\mathcal{M}_{NLQ \rightarrow SQL}(d_{i}, q_{i})
\end{equation}
Here $f(\cdot)$ denotes the schema retrieval function that selects the most relevant schema $d_i$ for question $q_i$. While effective to some extent, such two-stage methods still suffer from two major challenges. First, due to schema mismatch caused by semantically similar tables and columns in large database schemas $D$, the retrieved schema $d_i$ often deviates from the true target, leading to incorrect SQL generation. Second, the semantic gap between the natural language question $q_i$ and the corresponding SQL query $s_i$ can result in misinterpretation of user intent and generation of incorrect SQL.

To address these challenges, we propose CRED-SQL, as illustrated in figure~\ref{fig:cred-sql}. It comprises two key components:
\begin{enumerate}
\item Cluster-based Large-scale Schema Retrieval (CLSR): Mitigates schema mismatch by clustering attributes based on semantic similarity and down-weighting common (ambiguous) ones, enabling more accurate schema retrieval.
\item EDL-based SQL generation: A natural-language-based intermediate representation that explicitly describes SQL execution logic. By breaking the Text-to-SQL task into two subtasks—NLQ-to-EDL and EDL-to-SQL—we better align the semantic intent of the NLQ with the final SQL.
\end{enumerate}

The CRED-SQL pipeline proceeds as follows:
\begin{equation}
  d_{i} = f_{CLSR}(D, q_{i})
\end{equation}
\begin{equation}
\widetilde{e_{i}}=\mathcal{M}_{NLQ \rightarrow EDL}(d_{i}, q_{i})
\end{equation}
\begin{equation}
\widetilde{s_{i}}=\mathcal{M}_{EDL \rightarrow SQL}(d_{i}, q_{i}, \widetilde{e_{i}})
\end{equation}
where $f_{CLSR}$ denotes the cluster-based large-scale schema retrieval method, $\widetilde{e_{i}}$ represents the EDL generated by LLM $\mathcal{M}_{NLQ \rightarrow EDL}$, and $\widetilde{s_{i}}$ denotes the SQL generated by LLM $\mathcal{M}_{EDL \rightarrow SQL}$.

\subsection{Cluster-based Large-scale Schema Retrieve}

Schema retrieval is critical to the performance of LLM-based Text-to-SQL systems. To improve the identification of target schemas in large-scale databases with many semantically similar tables and columns, we propose CLSR, a cluster-aware schema retrieval framework. CLSR consists of three core components: (1) cluster-based schema indexing, (2) table retrieval, and (3) schema selection.

\textbf{Cluster-based Schema Indexing.} Traditional approaches typically represent a table by concatenating all its attributes into a single text sequence and embedding it as a whole. However, this strategy suffers from several drawbacks: (1) The concatenated representation often lacks linguistic coherence, diverging from natural text patterns. (2) It makes it difficult to distinguish important attributes, which degrades retrieval performance. 

To address these issues, we propose separating table information into two parts: table description and column information, and indexing them independently. Specifically, for each table, we encode its description using a pretrained embedding model and store the tuple \textit{<embedding, table name, table description>} in a vector database to form the table index.

Many tables contain attributes with similar semantics, making it difficult to distinguish between them. To mitigate this, we propose a column representation method based on semantic clustering. Each column $c_j$ is embedded as a vector $v_j \in \mathbb{R}^d$,  forming the full column vector set: $V = \{v_1, v_2, \dots, v_m\}$. We apply a clustering algorithm (e.g., K-means) to group semantically similar columns into $K$ clusters:
\begin{equation}
\mathcal{G}=\{G_1, G_2, ..., G_K\}
\end{equation}
Each cluster $G_k$ is represented by a centroid $\mu_k$, and each column is assigned to the nearest cluster:
\begin{equation}
Cluster(v_j) = arg \mathop{\min}_{k \in [1,K]} \lVert v_j-\mu_k \rVert ^{2}
\end{equation}
The cluster size $|G_k|$ reflects the frequency of similar columns. A larger cluster indicates high semantic redundancy and thus lower discriminative value; conversely, a smaller cluster suggests rarity and higher importance in distinguishing schemas. To handle large-scale schemas efficiently, we adopt a hybrid retrieval-based clustering algorithm (see Appendix A.2 \cite{duan2025credsqlenhancingrealworldlarge}). Given a similarity threshold \( s_1 \) and a predefined number of clusters $K$ the algorithm:
\begin{enumerate} [ topsep=-1pt, parsep=0pt]
\item Vectorizes all schema columns and retrieves the top-N most similar attributes using BM25 algorithm \cite{robertson2009probabilistic}. 

\item Assigns the current column to the most frequent cluster among the retrieved candidates, updating cluster sizes dynamically.
\end{enumerate}
Each column is ultimately represented as:
\begin{displaymath}
\textit{<embedding, cluster ID, cluster size, name, description, table>}
\end{displaymath}

\textbf{Table Retrieval.} We adopt a two-stage retrieval process. First, a set of candidate tables is retrieved using vector similarity against the table index. Then, we refine the ranking using a similarity scoring function that incorporates adaptive column weights:

\begin{equation}
  Score(T_{j}) = \sum_{i=1}^{n}Score(C_{ij})W_{i}+Score_{table}(T_{j}) 
\end{equation}
where $Score(C_{ij})$ denotes the similarity score  of the $i$-th column in table $T_j$, $W_{i}=\frac{1}{|G_k|_i}$ is the weight based on the column's cluster size,  $Score_{table}(T_{j})$ is the overall table similarity score from the initial retrieval stage. This formulation ensures that rare, informative attributes contribute more to table ranking, improving the precision of schema retrieval.

\textbf{Schema Selection.} Even after filtering, candidate tables may contain semantically similar attributes that confuse downstream SQL generation. To address this, we apply an LLM-based schema selection strategy that leverages in-context learning to refine both table and column choices. Following the framework of \cite{wang2023mac}, we design a schema selection prompt comprising four components: task description, selection instructions, candidate tables and columns, and few-shot examples. The LLM processes this prompt and outputs a refined subset of relevant tables and columns, then used for EDL generation.

\subsection{EDL-based SQL Generation} 
To mitigate the semantic gap between NLQ and SQL queries during the generation process, we introduce the SQL Execution Description Language (EDL)—an interpretable, structured, and hierarchical representation of query execution plans. To facilitate the training of both the Text-to-EDL and EDL-to-SQL modules, we construct two new datasets: Spider-EDL and Bird-EDL, based on the training and validation sets of the original Spider and Bird benchmarks.

\begin{table}[t]
\caption{Description of EDL Operators.}
\label{table:operator}
\centering
  \begin{tabularx}{\linewidth}{lX}

    \toprule
    \textbf{Operator} & \textbf{Format} \\
    \midrule
\makecell[l]{\textbf{Scan Table}} & \makecell[X]{Retrieve all rows from the [table name] table aliased as [alias].} \\
\midrule
\makecell[l]{\textbf{Join}} & \makecell[X]{Join the [table name] table aliased as [alias] on the condition that [condition].} \\
\midrule
\makecell[l]{\textbf{Reserve Rows}} & \makecell[X]{Reserve rows of [\#step\_number] where [filter condition].} \\
\midrule
\makecell[l]{\textbf{Subquery}} & \makecell[X]{Retrieve all rows from the [table name] table in a subquery and select the [column name] column. }\\
\midrule
\makecell[l]{\textbf{Group By}}  & \makecell[X]{Group [\#step\_number] by the [column name] column.} \\
\midrule
\makecell[l]{\textbf{Having Clause}} &\makecell[X]{Apply Having Clause: Reserve the grouped rows of [\#step\_number] where [condition].} \\
\midrule
\makecell[l]{\textbf{Sort}}  & \makecell[X]{Order [\#step\_number] by the [column name] column in [order type] order. }\\
\midrule
\makecell[l]{\textbf{Limit}} & \makecell[X]{Limit [\#step\_number] to the top [number] record(s). }\\
\midrule
\makecell[l]{\textbf{Select Column} } & \makecell[X]{Select the [column name] column from the [table alias] table in [\#step\_number].} \\
\midrule
\makecell[l]{\textbf{Set Operation} \\{\scriptsize(Union, Intersect, Except)}\\ } & \makecell[X]{Apply [set operation] operation: Exclude/Include the results in [query number] from/in the results in [query number].}\\
\midrule
\makecell[l]{\textbf{Arithmetic Calculation}} & \makecell[X]{Compute [column name] as the [arithmetic operation] of [column names or values].} \\
\midrule
\makecell[l]{\textbf{Date Calculation}} & \makecell[X]{Compute [column name] as the [date operation] of [column names or values].} \\
\midrule
\makecell[l]{\textbf{Case Statement}} & \makecell[X]{Compute [column name] as a case statement where [condition], then [result], else [default result].} \\
\midrule
\makecell[l]{\textbf{Substring Extraction}} & \makecell[X]{Extract substring from [column name] starting at position [start] for [length] characters as [column name].} \\
\midrule
\makecell[l]{\textbf{Casting}} & \makecell[X]{Cast [column name] as [new data type].} \\
\midrule
\makecell[l]{\textbf{Ranking} \\{\scriptsize(RANK OVER)}} & \makecell[X]{Compute the rank of [column name] ordered by [order column] in [order type] order using the RANK( ) window function.}  \\
    \bottomrule
  \end{tabularx}
\end{table}

\textbf{Execution Description Language.} Unlike QPL \cite{eyal2023semantic}, which adopts a SQL-like semi-structured format, EDL provides a more intuitive and human-readable representation of SQL execution plans in natural language. This approach offers two key advantages: (1) It enhances interpretability, especially for complex queries, by presenting execution logic in a step-by-step narrative form; (2) It supports a broader range of numerical operations through natural expressions, improving flexibility in describing analytical tasks. EDL structures each execution plan as a tree, where leaf nodes represent table scan operations and internal nodes represent logical operators such as joins, filtering, selections, and arithmetic computations. An EDL example from the Bird-EDL dataset is demonstrated in figure 3 in Appendix A.4 \cite{duan2025credsqlenhancingrealworldlarge}.

To generate EDL representations, we leverage GPT-4o to translate SQL queries into their corresponding EDL forms. To maintain consistency and expressiveness while simplifying the syntax, we define a standardized set of 16 core operators (see Table \ref{table:operator}). To ensure semantic equivalence between the original SQL and the generated EDL, we verify that both produce identical result sets on the same database, validating the accuracy and reliability of the conversion process.

\textbf{Text-to-EDL.} The Text-to-EDL module translates natural language questions into corresponding EDLs. This supervised learning task uses annotated examples where each question is paired with its ground-truth EDL. Each training instance in the Spider-EDL or Bird-EDL dataset is formatted as $\mathcal{X} = {(q_i, e_i, s_i, d_i)}$, where $q_i$ is the NLQ, $e_i$ is the EDL, $s_i$ is the corresponding SQL query, and $d_i$ is the associated database schema.

The model is trained to minimize the discrepancy between the predicted and gold EDLs:
\begin{equation}
  \mathop{\min}_{\mathcal{M}_{NLQ \rightarrow EDL}} \sum_{i=1}^{|\mathcal{X}|}\mathcal{L}_{\mathcal{M}_{NLQ \rightarrow EDL}}(\widetilde{e_{i}},e_{i})
\end{equation}
where $\mathcal{L}_{\mathcal{M}_{NLQ \rightarrow EDL}}$ is a loss function measuring the divergence between the generated EDL $\widetilde{e_{i}}$ and the ground truth $e_{i}$. The model learns to identify and align schema elements such as table names, columns, values, and operations from the question context, benefiting from the structured guidance of EDL.

\textbf{EDL-to-SQL.} The EDL-to-SQL module converts a structured EDL into a valid SQL query. This process is modeled as a structure-to-sequence generation task, where the input is a serialized or tree-structured EDL and the output is a SQL query.

The training objective is defined as:
\begin{equation}
  \mathop{\min}_{\mathcal{M}_{EDL \rightarrow SQL}} \sum_{i=1}^{|\mathcal{X}|}\mathcal{L}_{\mathcal{M}_{EDL \rightarrow SQL}}(\widetilde{s_{i}},s_{i})
\end{equation}
where $\mathcal{L}_{\mathcal{M}_{EDL \rightarrow SQL}}$ measures the difference between the generated query $\widetilde{s_{i}}$ and the reference SQL query $s_{i}$. The use of EDL as an intermediate representation constrains and guides the SQL generation process, significantly reducing the risk of producing semantically incorrect or irrelevant queries. This structured intermediate form enhances interpretability and improves generalization, especially for complex query logic.

%% file: 4.Experiment.tex
\section{Experiment}

\begin{table*}[t]
\belowrulesep=0pt
\aboverulesep=0pt
\centering
  \caption{Performance of different solutions on the dev set of SpiderUnion and BirdUnion dataset (EX, \%)}
  \label{tab:overall}
  \begin{tabular}{ll|ccccc|cccc}
    \toprule
    \multirow{2}{*}{\textbf{Method}} &\multirow{2}{*}{\textbf{Model}} &
    \multicolumn{5}{c|}{\textbf{SpiderUnion}} &
    \multicolumn{4}{c}{\textbf{BirdUnion}} \\
    & & easy & medium & hard & extra & ALL  & easy & moderate & challenging & ALL \\ 
     
    \hline

\multicolumn{11}{c}{NLQ$\rightarrow$SQL} \\
 \hline
CRUSH  &  GPT-4o &63.7 &57.2 &46.6 &28.3 &52.3 &58.38 &42.24 &33.79 &51.17\\
CRUSH  &  Qwen2.5-Coder-32B  &64.1 &56.3 &42.0 &30.1 &51.5 &53.73 &37.72 &34.48 &47.07 \\
CRUSH + DIN-SQL & GPT-4o &63.3 &57.8 &31.0 &13.3 &47.5 &56.11 &41.94 &33.33 &49.67\\
CRUSH + MAC-SQL & GPT-4o &61.3& 61.0&46.0 &31.9 &53.9 &56.43 &41.94 &30.56 &49.61 \\ 
CRUSH + DAIL-SQL& GPT-4o  & 64.9 & 55.6 & 40.2 & 24.1 & 50.2 &56.22 &39.44 &37.93 &49.41 \\

 \hline
\multicolumn{11}{c}{NLQ$\rightarrow$QPL$\rightarrow$SQL} \\
 \hline
CRUSH+QPL  &  GPT-4o & 63.3 & 56.1 & 43.1 & 32.5 & 51.8 &54.59 &32.97 &25.52 &45.31 \\
CRUSH+QPL  &  Qwen2.5-Coder-32B & 58.5 & 53.1 & 39.7 & 29.5 & 48.4 &49.73 &31.68 &28.28 &42.24\\
CRUSH + DIN-SQL +QPL & GPT-4o & 60.5 & 62.8 & 45.4 & 43.4 & 56.2 &56.65 &40.65 &31.25 &49.41\\
CRUSH + MAC-SQL +QPL & GPT-4o&60.5 & 60.5 & 40.2 & 28.9 & 52.0 &52.65 &35.13 &26.21 &44.85\\
CRUSH + DAIL-SQL +QPL & GPT-4o   & 62.5 & 52.7 & 41.4 & 27.7 & 49.1 &-- &-- &-- &--  \\
 \hline
\multicolumn{11}{c}{\textbf{Close-source LLM NLQ$\rightarrow$EDL$\rightarrow$SQL  (Ours)}} \\
 \hline

CRED-SQL&  GPT-4o & 75.8 & 72.0 & 65.5 & 54.8 & 69.1 &64.65 &48.49 &48.97 &58.28 \\
CRED-SQL + DIN-SQL & GPT-4o & 73.0 & 73.5 & 63.8 & 54.2 & 68.7 &67.35 &51.29 &\textbf{49.66} &60.82\\

CRED-SQL  + MAC-SQL & GPT-4o & 73.4 & 73.1 & 64.9 & 52.4 & 68.5 &\textbf{67.89} &55.17 &46.90 &62.06\\
CRED-SQL  + DAIL-SQL& GPT-4o & 74.2 & 74.0 & 63.8 & 47.6 & 68.1 &62.27 &45.26 &43.45 &55.35 \\
 \hline
\multicolumn{11}{c}{\textbf{Open-source LLM NLQ$\rightarrow$EDL$\rightarrow$SQL  (Ours)}} \\
 \hline

CRED-SQL&DeepSeek-Coder-6.7B&80.6&72.2&57.5&38.6&66.3&56.00&36.85&28.97&47.65\\
CRED-SQL&DeepSeek-Coder-33B&76.2&72.0&54.6&41.0&65.1&58.92&38.79&29.66&50.07\\
CRED-SQL&CodeLlama-7b&76.2&	65.0&	56.3&	39.8&	62.2&	52.65&	33.62&	22.07&	44.00\\
CRED-SQL&CodeLlama-13b&78.2&	69.7&	53.4	&38.6	&64.0&	56.22	&37.28	&28.97	&47.91\\
CRED-SQL&CodeLlama-34b&77.8&	69.3&	62.1	&43.4&	66.0&	57.41&	39.22&	27.59&	49.09\\

CRED-SQL&  Qwen2.5-Coder-32B & \textbf{80.6}&	\textbf{77.6}&	65.5	&59.6	&\textbf{73.4}&	63.46&	47.20&	37.93&	56.13\\
CRED-SQL + DIN-SQL&  Qwen2.5-Coder-32B &77.8&	77.6	&65.5&	60.8	&72.9&	63.46&	52.16&	43.45&	58.15\\
CRED-SQL  + MAC-SQL&  Qwen2.5-Coder-32B &77.8&	77.1&	\textbf{66.7}	&60.2&	72.8&	66.81&\textbf{59.91}&	47.59&	\textbf{62.91}\\
CRED-SQL + DAIL-SQL&  Qwen2.5-Coder-32B &77.4&	76.9	&64.4	&\textbf{62.7}&	72.6&	57.73&	44.40&	42.76	&52.28 \\

    \bottomrule
\end{tabular}
\end{table*}

\subsection{Datasets}
We test on the following two benchmarks that are designed by \cite{kothyari2023crush4sql}, because of the absence of any pre-existing large-schema benchmark.

\textbf{SpiderUnion:} This benchmark is based on Spider \cite{yu2018spider}, a widely recognized Text-to-SQL benchmark. In the original Spider dataset, each question $q_{i}$ is mapped to one of 166 database schemas. To create a more challenging scenario that simulates a large-scale schema environment, SpiderUnion combines all 166 databases into a single unified database by prefixing each table name with its corresponding database name. The resulting schema is substantially larger, consisting of 4502 columns distributed across 876 tables. For the evaluation in this paper, we use 1034 questions drawn from the Spider development set. Unlike the original Spider setup, where each question $q_{i}$ is linked to a specific database identifier from the set of 166 databases, our approach does not assume that the question is associated with any particular database ID. This setup presents a more generalized challenge for Text-to-SQL models, requiring them to handle a larger, more complex schema without prior knowledge of the relevant database context.

\textbf{BirdUnion:} Following a similar approach to SpiderUnion, BirdUnion is constructed from Bird \cite{li2024can}. The development set contains 11 databases, each encompassing an average of approximately 6.82 tables and 10.64 columns. To create a unified schema, each table name is prefixed with its corresponding database name. This unified schema comprises 798 columns across 75 tables. In this paper, we evaluate the method using 1534 questions sourced from the Bird development set.

\subsection{Metrics} 
For the Text-to-SQL task, we adopt the execution accuracy (EX) metric to ensure a fair comparison, following the approach of a prior study \cite{talaei2024chess}. Execution accuracy measures the performance by comparing the execution results of the generated SQL statement with those of the ground truth SQL query after execution. 

For schema retrieval within a large database schema, we use the \textit{Recall} metric, as proposed in a previous study \cite{kothyari2023crush4sql}.\textit{ Recall} is calculated for a selected schema set $R(q)$ and is defined as $\frac{|R(g)|}{|R(g) \cap R(q)|}$, where $R(g) \subset \mathcal{D}$ represents the gold retrieval set. To maintain fairness in subsequent experiments, we measure recall based solely on table names. This is because existing Text-to-SQL solutions typically select columns for SQL query generation after the relevant tables have been retrieved.

\subsection{Models and Baselines} 

We evaluate the generalizability of our approach using one close-source LLM \textbf{GPT-4o} \cite{hurst2024gpt} and six open-source code LLMs ranging from 6.7B to 34B. These include \textbf{CodeLlama-7B/13B/34B} \cite{touvron2023llama}, \textbf{Deepseek-Coder-6.7B/33B} \cite{guo2024Deepseek}, and \textbf{Qwen2.5-Coder-32B} \cite{hui2024qwen2}. During training, we used a learning rate of $5e^{-5}$ and performed two epochs of LoRA fine-tuning.

To perform a comprehensive end-to-end Text-to-SQL evaluation, we first employ the state-of-the-art schema retrieval method \textbf{CRUSH} \cite{kothyari2023crush4sql} to identify relevant tables from large-scale databases based on the input natural language questions. For the subsequent SQL generation stage, we evaluate our framework using three top-performing models from the Spider and Bird leaderboards: \textbf{DIN-SQL} \cite{pourreza2024din}, \textbf{MAC-SQL} \cite{wang2023mac}, and \textbf{DAIL-SQL} \cite{gao2024text}. Additionally, we include \textbf{QPL} \cite{eyal2023semantic} as a baseline for intermediate SQL-like representation methods.

\subsection{Overall Comparison}

In this subsection, we evaluate the performance of existing SOTA methods against our proposed CRED-SQL framework for end-to-end Text-to-SQL tasks in large-scale database scenarios. To ensure a fair comparison under complex schema settings, we first use CRUSH to retrieve the top-10 most relevant tables from each database schema. Subsequently, various Text-to-SQL approaches are applied to generate the final SQL queries based on these retrieved tables. To further explore the effectiveness of different intermediate representations, we divide the SQL generation module into two stages: NLQ$\rightarrow$QPL$\rightarrow$SQL and NLQ$\rightarrow$EDL$\rightarrow$SQL, respectively. For the SpiderUnion dataset, we retain the few-shot examples selected by each method (e.g., DIN-SQL, MAC-SQL, DAIL-SQL), replacing their original SQL with either QPL from the Spider-QPL dataset or EDL from the Spider-EDL dataset, before generating SQL using the respective frameworks. Due to the absence of a publicly available Bird-QPL dataset, we manually construct QPL examples for the BirdUnion dataset using GPT-4o to generate initial outputs based on published QPL rules, followed by human verification and refinement for a small number of few-shot samples. However, as DAIL-SQL requires full QPL annotations from the training set to build its retrieval-augmented few-shot pool, we are unable to conduct the NLQ$\rightarrow$QPL$\rightarrow$SQL experiments with DAIL-SQL on BirdUnion.

As shown in table~\ref{tab:overall}, our proposed CRED-SQL significantly outperforms all baseline methods on both the SpiderUnion and BirdUnion datasets. Specifically, CRED-SQL achieves 73.4\% execution accuracy (EX) on SpiderUnion using Qwen2.5-Coder-32B, and 62.91\% EX on BirdUnion when combined with CLSR and MAC-SQL using the same model. A notable performance gap exists between our approach and schema-retrieval-based methods such as CRUSH. For instance, CRUSH achieves an EX score approximately 21.9 percentage points lower than CRED-SQL on SpiderUnion using Qwen2.5-Coder-32B. Similarly, on BirdUnion, CRUSH underperforms CRED-SQL by 12.45 percentage points EX when used with MAC-SQL and GPT-4o. We attribute this to CRUSH's vulnerability to semantic interference during schema retrieval, where semantically similar column and table names lead to incorrect schema selection.

Furthermore, our experiments reveal that symbolic intermediate representations like QPL provide limited benefit for SQL generation with LLM. Across different frameworks (DIN-SQL, MAC-SQL, DAIL-SQL) and LLMs, the NLQ→QPL→SQL pipeline performs comparably to or slightly worse than direct NLQ→SQL generation. This can be attributed to QPL’s strict adherence to symbolic and structural fidelity, which constrains the natural language generation capabilities of LLMs. In contrast, EDL—the natural-language-based intermediate representation—demonstrates greater utility in assisting LLMs to better understand the semantics of the Text-to-SQL task, thereby improving SQL generation accuracy.

These findings validate the effectiveness of CRED-SQL in handling complex, real-world databases through structured component optimization, especially in managing schema complexity. Our framework exhibits strong scalability across different LLMs, achieving over 62\% ALL-scores on the SpiderUnion dataset with various backbones. Notably, the open-source Qwen2.5-Coder-32B model outperforms the closed-source GPT-4o in multiple settings, achieving 4.3\% higher EX on SpiderUnion and 0.85\% better performance on BirdUnion when paired with the MAC-SQL baseline. This suggests that with proper architectural design and component optimization, open-source models can rival or even surpass proprietary models in complex reasoning tasks such as Text-to-SQL.

\subsection{Effect of CLSR}

\begin{table}[b]
\caption{Table recalls on SpiderUnion Dev and BirdUnion Dev}
\centering
\vspace{1.5em}
\begin{subtable}[t!]{0.95\linewidth}
\caption{Table recalls on SpiderUnion Dev}
\vspace{-0.6em}
\begin{tabular}{l@{\hspace{2.5mm}}l@{\hspace{2.5mm}}c@{\hspace{2.5mm}}c@{\hspace{2.5mm}}c@{\hspace{2.5mm}}c@{\hspace{2.5mm}}c} 
\toprule
\multirow{2}{*}{\textbf{Model}} & \multirow{2}{*}{\textbf{Method}} & \multicolumn{5}{c}{\textbf{Number of recall tables}}\\
&&\textbf{@1} & \textbf{@3} & \textbf{@5}&\textbf{@10} & \textbf{@15} \\
\midrule
\multirow{2}{*}{GPT-4o} 
& CRUSH & 0.0851 & 0.3056 & 0.4758 & 0.6954 & 0.8308 \\
& CLSR  & \textbf{0.4023} & \textbf{0.7707} & \textbf{0.8095} & \textbf{0.8259} & 0.8259 \\
\midrule
Qwen2.5-& CRUSH & 0.1296 & 0.3182 & 0.4487 & 0.7166 & 0.8356 \\
Coder-32B& CLSR  & \textbf{0.4197} & \textbf{0.8124} & \textbf{0.8540} & \textbf{0.8762} & \textbf{0.8762} \\
\bottomrule
\end{tabular}
\end{subtable}

\vspace{1.5em}  
\begin{subtable}[b!]{0.95\linewidth}
\caption{Table recalls on BirdUnion Dev}
\vspace{-0.6em}
\begin{tabular}{l@{\hspace{2.5mm}}l@{\hspace{2.5mm}}c@{\hspace{2.5mm}}c@{\hspace{2.5mm}}c@{\hspace{2.5mm}}c@{\hspace{2.5mm}}c} 
\toprule
\multirow{2}{*}{\textbf{Model}} & \multirow{2}{*}{\textbf{Method}} & \multicolumn{5}{c}{\textbf{Number of recall tables}}\\
&&\textbf{@1} & \textbf{@3} & \textbf{@5}&\textbf{@10} & \textbf{@15} \\
\midrule
\multirow{2}{*}{GPT-4o} 
& CRUSH & 0.0932 & 0.3207 & 0.5228 & 0.7621 & 0.8598 \\
& CLSR  & \textbf{0.1714} & \textbf{0.7125} & \textbf{0.8514} & \textbf{0.9785} & \textbf{0.9863} \\
\midrule
Qwen2.5-& CRUSH & 0.0769 & 0.2705 & 0.4159 & 0.6571 & 0.8233 \\
Coder-32B& CLSR  & \textbf{0.1688} & \textbf{0.7093} & \textbf{0.8468} & \textbf{0.9713} & \textbf{0.9791} \\
\bottomrule
\end{tabular}
\end{subtable}
\label{table:recall}
\end{table}
Table \ref{table:recall} presents the table retrieval performance of our method compared to CRUSH on the GPT-4o and Qwen2.5-Coder-32B, measuring recall across the top 1 to 15 retrieved tables. Across nearly all settings, our approach—\textbf{CRED-SQL (CLSR)}—consistently outperforms the current state-of-the-art, \textbf{CRUSH}. Notably, CRED-SQL (CLSR) achieves convergence at a recall of 3, indicating that the target table is highly likely to appear within the top three retrieved candidates. In contrast, CRUSH shows a more gradual improvement, requiring a larger number of retrieved tables to reliably include the correct one. This trend suggests that CRUSH’s performance is adversely affected by the presence of numerous semantically similar tables. By contrast, CRED-SQL (CLSR) effectively reduces such interference through its cluster-based schema representation, resulting in more precise table retrieval.

\subsection{Effect of Execute Description Language(EDL)}

To assess the effectiveness of our EDL-based SQL generation framework, we conduct experiments on the Spider development set using three generation pipelines: (1) NLQ$\rightarrow$SQL, (2) NLQ$\rightarrow$QPL$\rightarrow$SQL, and (3) NLQ$\rightarrow$EDL$\rightarrow$SQL. Since there is no Bird-QPL dataset, this evaluation is limited to Spider. We first integrate QPL and EDL as intermediate representations into three baselines—DIN-SQL, MAC-SQL, and DAIL-SQL—by replacing their SQL generation modules, and compare execution accuracy (EX) across the three pipelines. Additionally, we assess performance using LLMs of varying sizes to evaluate the generalizability of each approach.To isolate the impact of schema retrieval, we also conduct experiments using gold schemas directly extracted from the gold SQL, ensuring a controlled evaluation of SQL generation quality.

Table \ref{table:spider_dev} summarizes the results. Across all models and settings, the NLQ$\rightarrow$EDL$\rightarrow$SQL pipeline consistently outperforms NLQ$\rightarrow$QPL$\rightarrow$SQL, and in most cases also surpasses the direct NLQ$\rightarrow$SQL approach. These results demonstrate that our EDL design not only enhances the interpretability of the generation process but also improves execution accuracy by providing a more structured and semantically rich intermediate representation.

\begin{table}[!]
\caption{EX results of NLQ$\rightarrow$SQL, NLQ$\rightarrow$QPL$\rightarrow$SQL and NLQ$\rightarrow$EDL$\rightarrow$SQL on the Spider dev dataset (\%)}
\label{table:spider_dev}
\centering
\begin{tabular}{l@{\hspace{2mm}}c@{\hspace{1.5mm}}c@{\hspace{1.5mm}}c} 
\toprule
\textbf{Method \& Model} &\makecell[c]{\textbf{NLQ}\\ \textbf{$\rightarrow$SQL}} &\makecell[c]{\textbf{NLQ$\rightarrow$QPL}\\ \textbf{$\rightarrow$SQL}} &\makecell[c]{\textbf{NLQ$\rightarrow$EDL}\\ \textbf{$\rightarrow$SQL}} \\
\midrule
\multicolumn{4}{c}{Origin database schema}\\
\midrule
DIN-SQL+GPT-4o&78.1&77.9&\textbf{83.3}\\
DIN-SQL+Qwen2.5-Coder-32B&81.6&79.5&\textbf{81.4}\\
MAC-SQL+GPT-4o&81.2&82.5&\textbf{83.1}\\
MAC-SQL+Qwen2.5-Coder-32B&78.4&78.0&\textbf{80.9}\\
DAIL-SQL+GPT-4o&84.4&76.1&\textbf{83.2}\\
DAIL-SQL+Qwen2.5-Coder-32B&80.2&74.9&\textbf{83.4}\\
GPT-4o&81.8&83.3&\textbf{83.5}\\
Qwen2.5-Coder-32B & 82.5 & 70.3 & \textbf{82.9} \\
DeepSeek-Coder-6.7B & 73.8 & 70.0 & \textbf{75.7} \\
DeepSeek-Coder-33B & 75.0 & 73.6 & \textbf{75.8} \\
CodeLlama-7b & 62.5 & 67.7 & \textbf{72.2} \\
CodeLlama-13b & 71.4 & 73.0 & \textbf{74.4} \\
CodeLlama-34b & 71.5 & 74.5 & \textbf{76.4} \\
\midrule
\multicolumn{4}{c}{Gold database schema}\\
\midrule
GPT-4o&83.0 & 84.0&\textbf{84.2} \\
Qwen2.5-Coder-32B & 80.9 & 75.9 & \textbf{82.7} \\
DeepSeek-Coder-6.7B & 79.5 & 73.7 & \textbf{79.7} \\
DeepSeek-Coder-33B & 78.4 & 75.0 & \textbf{80.9} \\
CodeLlama-7b & 68.6 & 72.1 & \textbf{79.4} \\
CodeLlama-13b & 73.1 & 77.0 & \textbf{79.0} \\
CodeLlama-34b & 78.9 & 78.4 & \textbf{81.3} \\

\bottomrule
\end{tabular}
\end{table}
Table \ref{table:gold_edl} presents the performance of GPT-4o and fine-tuned LLMs in converting EDL to SQL. For GPT-4o, we adopt a few-shot prompting approach using selected examples from the Spider-EDL training set. In contrast, other LLMs are fine-tuned directly on the same training data. All models are then evaluated on the Spider validation set using gold EDL as input. The results show that all models achieve execution accuracy (EX) above 98.4\%, confirming the strong semantic alignment between EDL and SQL and the reliability of EDL as an intermediate representation.

\begin{table}[!]
\caption{Results of EDL-to-SQL on the Spider dev dataset}
\label{table:gold_edl}
\centering
\begin{tabular}{ll} 
\toprule
\textbf{Model} & \textbf{EX(\%)}  \\
\midrule
GPT-4o & 99.5 \\
Qwen2.5-Coder-32B & 99.3\\
DeepSeek-Coder-6.7B&99.0\\
DeepSeek-Coder-33B&98.4\\
CodeLlama-7b&98.8\\
CodeLlama-13b&98.7\\
CodeLlama-34b&98.9\\

\bottomrule
\end{tabular}

\end{table}

\subsection{Abalation Study}

Table \ref{table:abalation_spider} presents the ablation results on the SpiderUnion development sets. Notably, due to the large number of database schemas, removing the CLSR component makes the experiment nearly infeasible. Therefore, in the ablation setting without CLSR, we use CRUSH as a substitute for schema retrieval. The results show that CLSR yields a substantial performance improvement—23.2\% in execution accuracy (EX) on the SpiderUnion development set. We also evaluated the impact of removing the EDL module. On the SpiderUnion development set, using EDL improved performance by 0.9 percentage points EX compared to direct SQL generation. 

\begin{table}[!]

\centering
\caption{Abalation Study on SpiderUnion Dev (EX, \%)}
\label{table:abalation_spider}
\begin{tabular}{lccccc} 
\toprule
\textbf{Pipeline} & \textbf{Easy} & \textbf{Medium} & \textbf{Hard}& \textbf{Extra} & \textbf{All} \\ 
\midrule
CRED-SQL &80.6 &77.6&65.5&59.6&73.4 \\
w/o CLSR& \multirow{2}{*}{61.3}& \multirow{2}{*}{55.2}& \multirow{2}{*}{40.2}& \multirow{2}{*}{30.7}& \multirow{2}{*}{50.2}\\
\textit{{\tiny(CRUSH+NLQ$\rightarrow$EDL$\rightarrow$SQL)}}&&&&& \\
w/o EDL& \multirow{2}{*}{81.9}& \multirow{2}{*}{76.5}& \multirow{2}{*}{63.8}& \multirow{2}{*}{57.2}& \multirow{2}{*}{72.5}\\
\textit{{\tiny(CLSR+NLQ$\rightarrow$SQL)}}&&&&& \\

\bottomrule

\end{tabular}
\end{table}

%% file: Conclusion.tex
\section{Limitations and Future Works}

All open-source LLMs mentioned in this paper were trained on NVIDIA A800 GPUs, and the version of close-source LLM, GPT-4o is gpt-4o-2024-08-06, whose input price is \$2.5 per 1M tokens and the output price is \$10 per 1M tokens. 

There are several limitations and areas for potential improvement in our current work. Due to our division of the Text-to-SQL process into two stages, Text-to-EDL and EDL-to-SQL, while the accuracy of SQL generation has improved, the response time for generating SQL has also increased. As illustrated in table 7 in Appendix A.1 \cite{duan2025credsqlenhancingrealworldlarge}, under the same model, the total duration of each NLQ$\rightarrow$EDL$\rightarrow$SQL on average is approximately three times that of NLQ$\rightarrow$SQL. Future research could focus on fine-tuning the LLM specifically for schema selection to enhance the precision in selecting the most relevant tables and columns. 

\section{Conclusion}

In this paper, we propose CRED-SQL, a novel framework that addresses two key challenges in Text-to-SQL systems for large-scale databases: schema mismatch and semantic deviation during SQL generation. By integrating Cluster-based Large-scale Schema Retrieval (CLSR) and Execution Description Language (EDL), CRED-SQL effectively bridges the semantic gap between natural language questions (NLQs) and SQL queries, offering a scalable and accurate solution for real-world applications.

%% file: Appendix.tex
\clearpage


\appendix
\section{Appendix}
\subsection{Limitations and Future Works}
\label{app:time}

All open-source LLMs mentioned in this paper were trained on NVIDIA A800 GPUs with 80GB RAM, and the version of close-source LLM, GPT-4o is gpt-4o-2024-08-06, which input price is \$2.5 per 1M tokens and the output price is \$10 per 1M tokens. Due to our division of the Text-to-SQL process into two stages—text-to-EDL and EDL-to-SQL—while the accuracy of SQL generation has improved, the response time for generating SQL has increased, as illustrated in the table \ref{table:spider_time_cost}.
\begin{table}[h!]
\caption{Average Response Time for different LLMs between NLQ$\rightarrow$SQL and NLQ$\rightarrow$EDL$\rightarrow$SQL on the Spider dev dataset (second per SQL)}
\label{table:spider_time_cost}
\centering
\begin{tabular}{lcc} 
\toprule
\textbf{Method \& Model} &\makecell[c]{\textbf{NLQ} \textbf{$\rightarrow$SQL}}  &\makecell[c]{\textbf{NLQ$\rightarrow$EDL} \textbf{$\rightarrow$SQL}} \\
\midrule
GPT-4o&1.2934&3.1357\\
Qwen2.5-Coder-32B&4.1712&14.4428\\
DeepSeek-Coder-6.7B&1.9216&5.3816\\
DeepSeek-Coder-33B&4.9588&14.0471 \\
CodeLlama-7b&2.2794&8.5319 \\
CodeLlama-13b&7.3956&18.8289 \\
CodeLlama-34b&5.2355&18.1131 \\
\bottomrule
\end{tabular}
\end{table}

There are several limitations and areas for potential improvement in our current work. First, during the schema selection process, we rely solely on the general reasoning capabilities of the LLM without applying any fine-tuning specific to this task. Future research could focus on fine-tuning the LLM specifically for schema selection to enhance the precision in selecting the most relevant tables and columns. 

Second, our approach involves manually constructing a Text-to-EDL dataset for fine-tuning the LLM to enhance the generation of Execute Description Language (EDL). While this method has proven effective, it is time-consuming because of two-stage SQL generation and it is also limited by the scope of manually curated data. Future research could explore more automated or semi-automated approaches for dataset construction, incorporating active learning techniques or leveraging data augmentation strategies to create larger and more diverse datasets. 

\subsection{Semantic Similarity Clustering Algorithm}
\label{app:Algorithm}
To efficiently cluster columns within large-scale database schemas, we introduce a hybrid retrieval-based clustering algorithm in CRSL. The main procedures, as outlined in Algorithm \ref{alg:cluster}, are as follows:

\begin{enumerate} 
\item All columns in the database schema are vectorized, and the top N attributes most relevant to the current attribute are retrieved using the BM25 algorithm \cite{robertson2009probabilistic}. 

\item A subset of attributes is selected based on a pre-set threshold. The current attribute is then classified into the cluster with the highest frequency within this set, and the cluster size is updated accordingly. 
\end{enumerate}

\begin{algorithm}[t]
\caption{Semantic Similarity Clustering Algorithm}
\label{alg:cluster}
\begin{algorithmic}[1]

\REQUIRE the similarity threshold $s_{1}$, number of clusters $n_{c}$
\ENSURE List of each column's unique identifier, vector, its cluster, and cluster size

\STATE $column\_vectors \gets \text{Vectorize}(columns)$ 
\STATE $current\_max\_cat \gets 0$ 

\STATE $visited\_vector \gets []$ 

\FOR{each $(uuid, v)$ in $column\_vectors$}
    \IF{$\text{len}(visited\_vector) == 0$}
        \STATE $visited\_vector.append(\{$
            \STATE \quad $\phantom{{}\{}'uuid': uuid,$
            \STATE \quad $\phantom{{}\{} 'vector': v,$
            \STATE \quad $\phantom{{}\{} 'cluster\_categories': 0,$
            \STATE \quad $\phantom{{}\{} 'cluster\_size': 1$
            \STATE \quad $\})$
            
    \ELSE
        \STATE $filtered\_results \gets []$
        \STATE $cluster\_categories\_list \gets [0] * n_{c}$ 
        \FOR{each $vd$ in $visited\_vector$}
            \STATE $similarity \gets \text{ComputeSimilarity}(v, vd['vector'])$
            \IF{$similarity > s_{1}$}
                \STATE $filtered\_results.append(vd)$
                \STATE Update the count of categories for $vd$
            \ENDIF
        \ENDFOR
        \IF{$\text{len}(filtered\_results) == 0$}
            \STATE $current\_max\_cat \mathrel{+}= 1$
            \STATE $visited\_vector.append(\{\hspace{1em}$ 
                
                \STATE \quad $\phantom{{}\{}'uuid': uuid,$
                \STATE \quad $\phantom{{}\{} 'vector': v,$
                \STATE \quad $\phantom{{}\{} 'cluster\_categories': current\_max\_cat,$
                \STATE \quad $\phantom{{}\{} 'cluster\_size': 1$
                \STATE \quad $\phantom{{}\{}\})$
        \ELSE
            \STATE $m \gets $ maximum index in  $cluster\_categories\_list$
            \FOR{each $item$ in $visited\_vector$}
                \IF{$item['cluster\_categories'] == max\_cluster\_cat$}
                    \STATE $item['cluster\_size'] \mathrel{+}= 1$
                    \STATE $clu\_size \gets item['cluster\_size']$
                \ENDIF
            \ENDFOR
            \STATE $visited\_vector.append(\{$
                \STATE \quad $\phantom{{}\{}'uuid':uuid,$
                \STATE \quad $\phantom{{}\{} 'vector':v, $
                \STATE \quad $\phantom{{}\{}'cluster\_categories':m, $
                \STATE \quad $\phantom{{}\{}'cluster\_size':c\_s$
                \STATE \quad $\phantom{{}\{}\})$
        \ENDIF
    \ENDIF
\ENDFOR

\RETURN $visited\_vector$

\end{algorithmic}
\end{algorithm}

\subsection{More Results}
\subsubsection{More results of CLSR on BirdUnion dev dataset}
Table \ref{table:CLSR_K} presents table recall results on BirdUnion Dev with different similarity thresholds. As shown in Algorithm~\ref{alg:cluster} in the appendix~\ref{app:Algorithm}, to prevent the impact of an inappropriate preset number of clusters on the results, the number of clusters in our algorithm is determined by a similarity threshold. Specifically, when the similarity between an attribute and existing clusters falls below a certain threshold, a new cluster is created; otherwise, the attribute is assigned to the most similar cluster. This approach only requires setting a very low initial number of clusters (e.g., K=1), with the number of clusters updated dynamically during computation. The results in table~\ref{table:CLSR_K} show that as the threshold increases from 0.4 to 0.8, the number of clusters (K) remains consistently at 620, and the recall on tables remains nearly unchanged. This demonstrates the robustness of our method and its insensitivity to the number of clusters.

\begin{table}[!]

\centering
\caption{Table recalls on BirdUnion Dev with different similarity thresholds}
\label{table:CLSR_K}
\begin{tabular}{cccccccc} 
\toprule
\textbf{Similarity}   & \multirow{2}{*}{\textbf{K}} & \multicolumn{5}{c}{\textbf{Number of recall tables}}\\
\textbf{threshold}&&\textbf{@1} & \textbf{@3} & \textbf{@5}&\textbf{@10} & \textbf{@15} \\
\midrule

0.8 &620&0.1467 & 0.7080 & 0.8520 & 0.9772 & 0.9876 \\
0.7 &620&0.1558 & 0.7080 & 0.8533 & 0.9765 & 0.9870 \\
0.6 &620&0.1578 & 0.7093 & 0.8520 & 0.9798 & 0.9863\\
0.5 &620&0.1610 & 0.7268 & 0.8585 & 0.9759 & 0.9857\\
0.4 &620&0.1688 & 0.7093 & 0.8468 & 0.9713 & 0.9791\\

\bottomrule

\end{tabular}
\end{table}

\subsubsection{Few-shot results on origin schema Spider dev dataset}
Table \ref{table:spider_fewshot} presents results of different LLMs on the Spider development set using three generation pipelines: (1) NLQ$\rightarrow$SQL, (2) NLQ$\rightarrow$QPL$\rightarrow$SQL, and (3) NLQ$\rightarrow$EDL$\rightarrow$SQL. Unlike table \ref{table:spider_dev}, here we adopt the few-shot approach for LLM instead of fine-tuning. The results show that across nearly all models and settings, the NLQ$\rightarrow$EDL$\rightarrow$ SQL pipeline consistently outperforms NLQ$\rightarrow$QPL$\rightarrow$SQL, and in GPT-4o also surpasses the direct NLQ$\rightarrow$SQL approach. Among other models, NLQ$\rightarrow$EDL$\rightarrow$ SQL is inferior to NLQ$\rightarrow$SQL. This is because LLMs with smaller sizes have insufficient learning ability from few-shot examples compared to fine-tune.

\begin{table}[H]
\caption{EX results of few-shot NLQ$\rightarrow$EDL$\rightarrow$SQL, NLQ $\rightarrow$ QPL $\rightarrow$ SQL and NLQ $\rightarrow$ SQL on the origin schema Spider dev dataset (\%)}
\label{table:spider_fewshot}
\centering
\begin{tabular}{lccc} 
\toprule
\textbf{Model} &\makecell[c]{\textbf{NLQ}\\ \textbf{$\rightarrow$SQL}} &\makecell[c]{\textbf{NLQ$\rightarrow$QPL}\\ \textbf{$\rightarrow$SQL}} &\makecell[c]{\textbf{NLQ$\rightarrow$EDL}\\ \textbf{$\rightarrow$SQL}} \\
\midrule
GPT-4o & 81.8&83.3& \textbf{83.5}\\
Qwen2.5-Coder-32B &83.8 &78.5& 82.8\\
DeepSeek-Coder-6.7B & 70.3&59.9 &63.4\\
DeepSeek-Coder-33B & 75.2 &68.5&63.2\\
CodeLlama-7b & 54.6&29.4& 36.4\\
CodeLlama-13b & 57.8&43.9& 45.2\\
CodeLlama-34b & 70.9&28.7& 58.7\\

\bottomrule
\end{tabular}
\end{table}

\subsubsection{Other results on Bird dev dataset}

As a supplement, we also conduct experiments on the Bird development set using three generation pipelines: (1) NLQ$\rightarrow$SQL, (2) NLQ $\rightarrow$QPL$\rightarrow$SQL, and (3) NLQ$\rightarrow$EDL$\rightarrow$SQL. Since there is no Bird-QPL dataset, we only generated more than ten QPLs required for DIN-SQL, MAC-SQL and few-shot LLMs using GPT-4o based on the rules of QPL and some Spider-QPLs as few-shots, and manually checked and modified them. DAIL-SQL method cannot apply QPL because there is no bird-QPL training set as the retrieval data set.

\begin{table}[!]
\caption{EX results of NLQ$\rightarrow$SQL, NLQ$\rightarrow$QPL$\rightarrow$SQL and NLQ$\rightarrow$EDL$\rightarrow$SQL on the Bird dev dataset (\%)}
\label{tab:bird_dev}
\centering
\begin{tabular}{l@{\hspace{2mm}}c@{\hspace{1.5mm}}c@{\hspace{1.5mm}}c} 
\toprule
\textbf{Method \& Model} &\makecell[c]{\textbf{NLQ}\\ \textbf{$\rightarrow$SQL}} &\makecell[c]{\textbf{NLQ$\rightarrow$QPL}\\ \textbf{$\rightarrow$SQL}} &\makecell[c]{\textbf{NLQ$\rightarrow$EDL}\\ \textbf{$\rightarrow$SQL}} \\
\midrule
DIN-SQL+GPT-4o&60.23&57.17&\textbf{61.86}\\
DIN-SQL+Qwen2.5-Coder-32B&58.15&54.30&\textbf{59.58}\\
MAC-SQL+GPT-4o&60.76&60.56&\textbf{63.04}\\
MAC-SQL+Qwen2.5-Coder-32B&61.80&62.78&\textbf{64.47}\\
DAIL-SQL+GPT-4o&54.74&--&\textbf{56.32}\\
DAIL-SQL+Qwen2.5-Coder-32B&48.17&--&\textbf{53.72}\\
GPT-4o&62.06&56.58&59.32\\
Qwen2.5-Coder-32B & 56.91 & 55.15 & 58.34 \\
DeepSeek-Coder-6.7B & 36.64 & 30.96 & 32.86 \\
DeepSeek-Coder-33B & 44.46 & 42.63 & 38.92 \\
CodeLlama-7b & 23.53 & 12.39 & 15.45 \\
CodeLlama-13b & 26.86 & 17.47 & 20.34 \\
CodeLlama-34b & 33.38 & 24.97 & 28.49 \\
\bottomrule
\end{tabular}
\end{table}

We first integrate QPL and EDL as intermediate representations into three baselines—DIN-SQL, MAC-SQL, and DAIL-SQL—by replacing their SQL generation modules, and compare execution accuracy (EX) across the three pipelines. Additionally, we assess performance using few-shot LLMs of varying sizes to evaluate the generalizability of each approach. As shown in table \ref{tab:bird_dev}, across all baselines with GPT-4o and Qwen2.5-Coder-32B, the NLQ $\rightarrow$ EDL $\rightarrow$ SQL pipeline consistently outperforms NLQ $\rightarrow$ QPL $\rightarrow$ SQL, and also surpasses the direct NLQ $\rightarrow$ SQL approach. Across nearly all few-shot LLMs, the NLQ $\rightarrow$ EDL $\rightarrow$ SQL pipeline consistently outperforms NLQ $\rightarrow$ QPL $\rightarrow$ SQL, but is inferior to NLQ $\rightarrow$ SQL. This is because LLMs with smaller sizes have insufficient learning ability from few-shot examples compared to fine-tune.

Table \ref{tab:bird_dev_finetune} presents the NLQ $\rightarrow$ EDL $\rightarrow$ SQL and NLQ $\rightarrow$ SQL performance of fine-tuned LLMs on the Bird dev dataset. To isolate the impact of schema retrieval, we also conduct experiments using gold schemas directly extracted from the gold SQL, ensuring a controlled evaluation of SQL generation quality. Results show that when the size of LLM parameters is small, such as DeepSeek-Coder-6.7B and CodeLlama-7B , EDL is more helpful for SQL generation, but its performance on LLMs with a larger number of parameters does not exceed the performance of NLQ$\rightarrow$SQL. This is because SQL queries in the Bird dataset are usually more complex than those in the Spider dataset, and evidence is introduced to illustrate complex queries, which pose a greater challenge for LLMs to generate correct EDLs.

\begin{table}[!]
\caption{EX results of NLQ$\rightarrow$SQL and NLQ$\rightarrow$EDL$\rightarrow$SQL with different fine-tuned LLMs on the Bird dev dataset (\%)}
\label{tab:bird_dev_finetune}
\centering
\begin{tabular}{lcc} 
\toprule
\textbf{Method \& Model}  &\makecell[c]{\textbf{NLQ$\rightarrow$SQL}} &\makecell[c]{\textbf{NLQ$\rightarrow$EDL} \textbf{$\rightarrow$SQL}} \\
\midrule
\multicolumn{3}{c}{Origin database schema}\\
\midrule
Qwen2.5-Coder-32B & 58.41  & 53.19 \\
DeepSeek-Coder-6.7B & 36.11 & \textbf{48.70} \\
DeepSeek-Coder-33B & 51.89  & 48.31 \\
CodeLlama-7b &37.48  & \textbf{42.57} \\
CodeLlama-13b &43.61  & \textbf{45.05} \\
CodeLlama-34b & 48.50 & 48.17 \\
\midrule
\multicolumn{3}{c}{Gold database schema}\\
\midrule
Qwen2.5-Coder-32B & 64.60 &  61.02 \\
DeepSeek-Coder-6.7B & 47.72  & \textbf{52.15} \\
DeepSeek-Coder-33B & 57.30 &  55.22 \\
CodeLlama-7b & 47.39  & \textbf{49.61} \\
CodeLlama-13b & 53.06  & 51.50\\
CodeLlama-34b & 55.28 & 52.09 \\

\bottomrule
\end{tabular}
\end{table}

\begin{table}[!]
\caption{Gold EDL to SQL results on Bird dev Dataset}
\centering
\begin{tabular}{ll} 
\toprule
\textbf{Model} & \textbf{EX(\%)}  \\
\midrule
GPT-4o & 96.61 \\
Qwen2.5-Coder-32B & 93.55\\
DeepSeek-Coder-6.7B&92.18\\
DeepSeek-Coder-33B&91.26\\
CodeLlama-7b&89.37\\
CodeLlama-13b&91.07\\
CodeLlama-34b&91.66\\

\bottomrule
\end{tabular}
\label{table:gold_edl_bird}
\end{table}

Table \ref{table:gold_edl_bird} presents that although the EX results of the LLMs from gold EDL to SQL are around 90\% or above, it does not reach the 99\% EX in Spider dataset in table \ref{table:gold_edl}. Therfore, future research could explore more automated or semi-automated approaches for dataset construction, incorporating leveraging data augmentation strategies to create larger and more diverse datasets to improve the quality of EDL on Bird dataset, thereby enhancing the accuracy of SQL generation.

\subsection{Case Study}
\label{app:case_study}
\begin{figure}[!]
  \centering
  \includegraphics[width=0.9\linewidth]{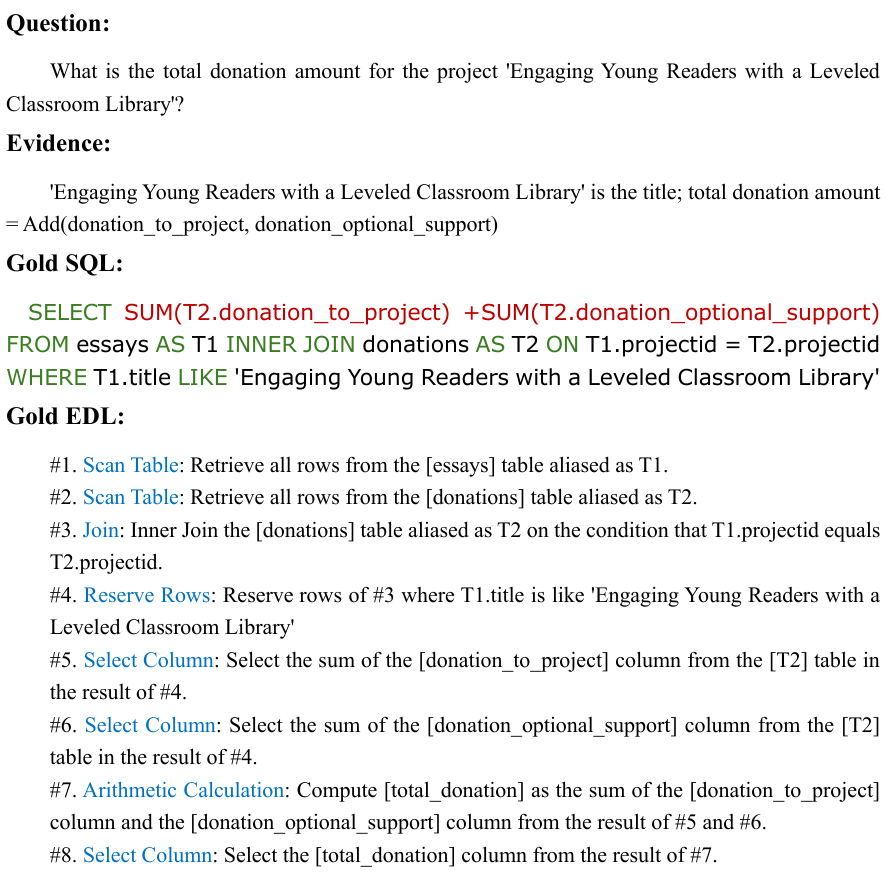}
  \caption{Example EDL on the Bird-EDL dataset.}
  \label{fig:edl}
  \vspace{0.5cm}
  
\end{figure}

Figure~\ref{fig:edl} illustrates an example from the Bird-EDL dataset comprising eight steps and five unique operators, including two \textit{Scan Table} steps, followed by a \textit{Join}, \textit{Reserve Rows}, and three \textit{Select Column} operations. The final step involves an \textit{Arithmetic Calculation} operator applied to previously selected columns. This hierarchical structure effectively captures both data flow and logical sequence of operations.

We conduct the case analysis of the results from the Qwen2.5-Coder-32B on the SpiderUnion dataset, as shown in table \ref{tab:case}. Example 1 presents a case analysis for the schema mismatch error. The NLQ is: \textit{"Find the number of cities in each district whose population is greater than the average population of cities?" } The correct schema is the \texttt{city} table in \texttt{world\_1} database. However, due to the presence of numerous table and column names in the database that are semantically similar to keywords such as \textit{"district", "population"} from the question, CRUSH with Qwen2.5-Coder-32B ranks other tables as more relevant and the target table is not included among the top ten retrieved tables. This led to the subsequent error of NLQ$\rightarrow$EDL $\rightarrow$SQL, which selected the wrong table \texttt{district} and columns (\texttt{Headquartered\_City, District\_name}, \texttt{ Regional\_Population}).

Example 2 presents a case analysis for semantic deviation in SQL generation, the NLQ is: \textit{"Find the major and age of students who do not have a cat pet." }The correct schema is the \texttt{Student,Has\_Pet} and \texttt{Pets} tables in the \texttt{pets\_1} database. In this example, we first conduct schema retrieval using our CRED-SQL (CLSR), and the result shows that three target tables are included in the top three retrieved candidates. During the SQL generation stage, the SQL query is directly generated from NLQ and the retrieved schema by Qwen2.5-Coder-32B exists semantic deviation. In this SQL statement, the \texttt{EXCEPT} difference is based on the (major, age) combination. If multiple students share the same (major, age), then as long as one of them owns a cat, that (major, age) group will be excluded. This may result in multiple students being either excluded or retained, which differs from the actual intended behavior of NLQ. While EDL corrected this semantic deviation using \texttt{NOT IN} instead, as shown in CRED-SQL result. This SQL query generated from EDL uses \texttt{StuID} as the unique identifier to exclude specific students and make the query results conform to the NLQ search intention.

\begin{table}[!]
\centering
  \caption{Case Analysis of Qwen2.5-Coder-32B on SpiderUnion Dev}
  \label{tab:case}
  \begin{tabularx}{\linewidth}{X}
    \toprule
    \textbf{Example 1: Schema mismatch error} \\
    \midrule
\makecell[X]{\textit{Find the number of cities in each district whose population is greater than the average population of cities?}} \\
\sethlcolor{yellow}
\hl{\textbf{Gold database:}} world\_1\\

\hl{\textbf{Gold SQL:}} \sethlcolor{green}SELECT count(*) ,  \hl{District} FROM 
\hl{city} WHERE \hl{Population}  >  (SELECT avg(\hl{Population}) FROM \hl{city}) GROUP BY \hl{District}\\
\hdashline
\sethlcolor{yellow}
\hl{\textbf{CRUSH schema retrieval result:}}\\
\makecell[X]{"city\_record.city",\\ 
"county\_public\_safety.county\_public\_safety", \\
"farm.city", \\
"geo.city", \\
"store\_product.district", \\
"company\_office.buildings", \\
"geo.state", \\
"debate.people", \\
"city\_record.temperature", \\
"world\_1.country"}\\
\hl{\textbf{CLSR schema retrieval result:}}\\
\makecell[X]{\sethlcolor{green}\hl{"world\_1.city",}\\
"world\_1.sqlite\_sequence",\\
"world\_1.country",\\
"world\_1.countrylanguage"
}\\
\hdashline
\sethlcolor{yellow}
\hl{\textbf{CRUSH+NLQ$\rightarrow$EDL$\rightarrow$SQL result}} \\
\sethlcolor{pink}
\texttt{SELECT count(*) , T1.\hl{Headquartered\_City} FROM \hl{district} AS T1 JOIN city AS T2 ON T1.\hl{District\_name} = T2.city WHERE T2.\hl{Regional\_Population} > (SELECT avg(\hl{regional\_population}) FROM city) GROUP BY T1.\hl{Headquartered\_City}}
 \makecell[X]{} \\
\sethlcolor{yellow}

\hl{\textbf{CRED-SQL result}} \\
\sethlcolor{green}
\texttt{SELECT count(*) , \hl{District} FROM \hl{city} AS T1 WHERE \hl{Population} > (SELECT avg(\hl{Population}) FROM \hl{city}) GROUP BY \hl{District}}\\
\midrule

 
\textbf{Example 2: Semantic deviation in SQL generation. } \\
\midrule
\textit{Find the major and age of students who do not have a cat pet.}\\
\hl{\textbf{Gold database:}} pets\_1\\

\hl{\textbf{Gold SQL:}} \sethlcolor{green}SELECT major ,  age FROM student WHERE stuid \hl{NOT IN} (SELECT T1.stuid FROM student AS T1 JOIN has\_pet AS T2 ON T1.stuid  =  T2.stuid JOIN pets AS T3 ON T3.petid  =  T2.petid WHERE T3.pettype  =  'cat')\\
\hdashline

\hl{\textbf{CLSR schema retrieval result:}}\\
\sethlcolor{green}
\makecell[X]{\hl{"pets\_1.Student",}\\
\hl{"pets\_1.Has\_Pet",}\\
\hl{"pets\_1.Pets"}
}\\
\hdashline
\sethlcolor{yellow}
\hl{\textbf{CLSR+NLQ$\rightarrow$SQL result}} \\
\sethlcolor{pink}
\texttt{SELECT major, age FROM Student \hl{EXCEPT} SELECT T1.major, T1.age FROM Student AS T1 JOIN Has\_Pet AS T2 ON T1.StuID = T2.StuID JOIN Pets AS T3 ON T2.PetID = T3.PetID WHERE T3.PetType = "cat"}\\
\sethlcolor{yellow}

\hl{\textbf{CRED-SQL result}} \\
\sethlcolor{green}
\makecell[X]{\textbf{Generated EDL: }\\\#1.Scan Table: Retrieve all rows from the [Student] table. \\\#2.Subquery: Retrieve all rows from the [Has\_Pet] table aliased as T1.\\\#3.Join the [Pets] table aliased as T2 on the condition that T1.PetID equals T2.PetID. \\\#4. Reserve rows of \#3 where the [PetType] in table T2 is 'cat'. \\\#5. Select the [StuID] column from the [T1] table from the result of \#4. \\\#6. Reserve rows of \#1 where [StuID] is \hl{not in} the result of \#5.\\ \#7. Select the [major] and [age] columns from the [Student] table from the result of \#6.}\\
\textbf{Generated SQL: }\\
\sethlcolor{green}
\texttt{SELECT major, age FROM Student WHERE StuID \hl{NOT IN} (SELECT T1.StuID FROM Has\_Pet AS T1 JOIN Pets AS T2 ON T1.PetID = T2.PetID WHERE PetType = 'cat')}\\
    \bottomrule
  \end{tabularx}
\end{table}